\documentclass[lettersize,journal,twoside]{IEEEtran}
\usepackage{amsmath,amsfonts}
\usepackage{algorithmic}
\usepackage{algorithm}
\usepackage{array}
\usepackage{textcomp}
\usepackage{stfloats}
\usepackage{url}
\usepackage{verbatim}
\usepackage{graphicx}
\usepackage{cite}
\usepackage{import}
\usepackage{pgfplots}
\pgfplotsset{compat=1.9, table/search path={data}}
\usepgfplotslibrary{fillbetween}
\usepackage{tabularray}
\usepackage{tabularx}
\usepackage{multirow}
\usepackage{colortbl}
\usepackage{booktabs}
\usepackage{siunitx}
\usepackage{amssymb}
\usepackage{subcaption}
\usepackage{textcomp,gensymb}
\usepackage[hidelinks]{hyperref}
\usepackage[nameinlink,capitalise]{cleveref}
\usepackage{url}
\usepackage{orcidlink}

\usepackage{acronym}
\acrodef{SLAM}{Simultaneous Localization and Mapping}
\definecolor{Black}{gray}{0.0}
\definecolor{Purple}{rgb}{0.77,0.12,0.64}
\definecolor{FullBlue}{rgb}{0,0,1}

\begin{document}
\title{BeautyMap: Binary-Encoded Adaptable Ground Matrix for Dynamic Points Removal in Global Maps}

\author{
    Mingkai~Jia\orcidlink{0000-0003-2100-5305},
    Qingwen~Zhang\orcidlink{0000-0002-7882-948X},
    Bowen Yang\orcidlink{0000-0002-3727-1248},
    ~\IEEEmembership{Graduate Student Member, IEEE}, \\
    Jin Wu\orcidlink{0000-0001-5930-4170},
    ~\IEEEmembership{Member,~IEEE},
    Ming Liu\orcidlink{0000-0002-4500-238X},~\IEEEmembership{Senior Member,~IEEE}, 
    and~Patric~Jensfelt\orcidlink{0000-0002-1170-7162},~\IEEEmembership{Member,~IEEE}
\thanks{
Manuscript received December 31, 2023; revised March 29, 2024; accepted April 29, 2024. 
This paper was recommended for publication by Editor Markus Vincze upon evaluation of the Associate Editor and Reviewers’ comments. 
This work was supported by Wallenberg AI, Autonomous Systems and Software Program (WASP). 
\textit{(Mingkai~Jia and Qingwen~Zhang are co-first authors.) (Corresponding author: Mingkai~Jia and Qingwen~Zhang.)}
}
\thanks{MingKai~Jia, Bowen Yang, and Jin Wu are with Robotics Institute, The Hong Kong University of Science and Technology, Hong Kong SAR, China. (email: mjiaab@connect.ust.hk)}
\thanks{Ming Liu is with The Hong Kong University of Science and Technology (Guangzhou), Nansha, Guangzhou, 511400, Guangdong, China.}
\thanks{Qingwen~Zhang and Patric~Jensfelt are with the Division of Robotics, Perception, and Learning (RPL), KTH Royal Institute of Technology, Stockholm 114 28, Sweden. (email: qingwen@kth.se; patric@kth.se)}
\thanks{Digital Object Identifier (DOI): see top of this page.}
}

\markboth{IEEE ROBOTICS AND AUTOMATION LETTERS. PREPRINT VERSION. ACCEPTED April~2024}%
{M. Jia, Q. Zhang, et. al: BeautyMap}


\maketitle

\begin{abstract}
Global point clouds that correctly represent the static environment features can facilitate accurate localization and robust path planning.
However, dynamic objects introduce undesired \textit{`ghost'} tracks that are mixed up with the static environment.
Existing dynamic removal methods normally fail to balance the performance in computational efficiency and accuracy. 
In response, we present \textit{`BeautyMap'} to efficiently remove the dynamic points while retaining static features for high-fidelity global maps.
Our approach utilizes a binary-encoded matrix to efficiently extract the environment features. 
With a bit-wise comparison between matrices of each frame and the corresponding map region, we can extract potential dynamic regions. Then we use coarse to fine hierarchical segmentation of the $z$-axis to handle terrain variations.
The final static restoration module accounts for the range-visibility of each single scan and protects static points out of sight.
Comparative experiments underscore BeautyMap's superior performance in both accuracy and efficiency against other dynamic points removal methods. 
The code is open-sourced at \href{https://github.com/MKJia/BeautyMap}{https://github.com/MKJia/BeautyMap}.
\end{abstract}


\section{Introduction}
\begin{figure}[t]
  \centering
  \includegraphics[width=\linewidth]{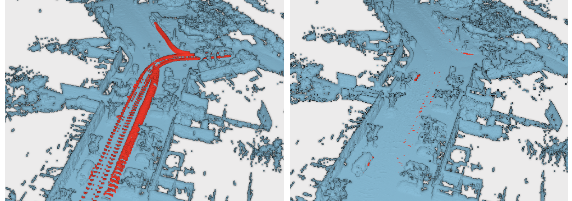}
  \caption{The results of dynamic points removal in SemanticKITTI dataset, utilizing our proposed approach called BeautyMap. \textbf{Red:} all of the dynamic points (left) and after removed (right). \textbf{Blue:} static points.}
  \label{fig:result}
\end{figure}

Point clouds are widely used in various robotics applications due to their effectiveness in supporting essential functions and tasks~\cite{jiao2022fusionportable, nguyen2022ntu, liu2021role, mikielevation2022}.
Existing \ac{SLAM} approaches \cite{TMSLICT, xu2022fast} simply fuse point clouds without intentionally removing dynamic points that may affect the accurate representation of static environment features.
Including dynamic points in the map can result in ghost points, as illustrated in red in~\cref{fig:result}. 
These ghost points will also affect the performance of downstream tasks. In the localization task, ghost points may reduce robustness. For global path planning, the presence of ghost points can lead to suboptimal path selection. 

Existing methods of dynamic point removal can be grouped into three categories based on their logic of pipeline: freespace-based, differential-based, and data-driven.
Freespace-based methods \cite{hornung2013octomap, oleynikova2017voxblox, duberg2020ufomap} identify free regions if they are frequently passed through during ray-casting.
However, the partial detection problem~\cite{zhang2023dynamic} and the high computation demands limit their applications.
Differential-based approaches \cite{kim2020remove, lim2021erasor} classify dynamic points by comparing differences between scans and maps. 
However, their lack of a suitable global map representation complicates the restoration of mistakenly removed static points. 
Data-driven methods~\cite{pfreundschuh2021dynamic,milioto2019rangenet++,wilson2022motionsc,mersch2023ral,zhang2024deflow,cheng2024mf} rely on learning from dataset and train a feature extraction network to detect dynamic points directly. 
Although these methods effectively identify moving objects, domain adaptation is a big challenge if the train and test datasets are from different sensor setups.

In this paper, we introduce our novel dynamic points removal method for open-world environments, referred to as the Binary-Encoded Adaptable groUnd maTrix for dYnamic points removal (BeautyMap). 
The proposed method can remove dynamic points from a global point cloud map efficiently and accurately. 
To enhance the efficiency, we represent point cloud data as 3D binary grids and encode them to 2D matrices. 
We mark all potential dynamic points by comparing matrices between frames and the global point cloud map using bitwise operations. 
This straightforward operation allows us to perform potential dynamic region detection more efficiently than traditional ray-casting methods. 

For accurate dynamic points removal, we perform an adaptable ground adjustment process.
In this process, we apply a coarse ground extraction module to filter outlier points below ground and a hierarchical resolution module to segment the upper surface of the ground and the dynamic points near the ground.
Then we perform a static points restoration process to review the former potential dynamic regions, where we consider the sensor property and mask out of sight detections. 
We evaluate our method through several datasets and BeautyMap achieves state-of-art performance on the dynamic removal in point cloud maps benchmark~\cite{zhang2023dynamic}. 
Our approach is open-source at \color{blue}\href{https://github.com/MKJia/BeautyMap}{https://github.com/MKJia/BeautyMap}\color{black}. 

\begin{figure*}[t!]
  \centering
  \includegraphics[width=\linewidth]{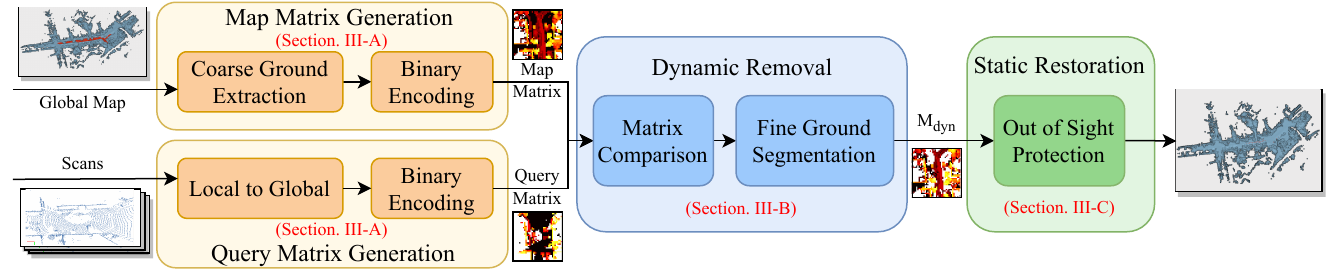}
  \caption{The overview pipeline of our proposed method. The global map is encoded as a matrix, and processed by the adaptable ground adjustment. Scans are similarly encoded after being transformed into the global frame. To realize dynamic region recognition, the query matrix is compared with the corresponding map submatrix bitwise to get a potential dynamic matrix. Finally, incorrectly classified points that are actually static are restored. }
  \label{fig:overview}
\end{figure*}
Our main contributions are as follows:
\begin{itemize}
\item We represent a binary 3D grid map as a binary-encoded matrix to improve computational efficiency through bit manipulation and matrix operations.
\item We introduce adaptive adjustment and hierarchical z-axis resolution for ground-level points and present a protection module for falsely detected points.
\item We validate BeautyMap against state-of-the-art methods and demonstrate its superior performance through a comprehensive evaluation.
\end{itemize}

\section{Related Works}
This section reviews dynamic point removal methods, categorized into three types: freespace-based, differential-based, and data-driven. 
We primarily focus on the map representation structures of these methods and their different approaches to achieving map fidelity and efficiency.

\subsection{FreeSpace-based Removal Algorithms}

Most freespace-based removal algorithms utilize occupancy voxel maps. 
Various data structures and their extensions aim to compress memory and enhance efficiency for a faster traverse of all voxels. 
Octomap \cite{hornung2013octomap} uses ray casting, marking voxels with points in them as occupied, voxels along the ray as free, and the rest remains unknown. 
Its Octree structure offers hierarchical resolution by equally dividing a cube into eight segments. 
Duberg \textit{et al.} \cite{duberg2020ufomap,daniel2024dufomap} expand the structure by explicitly storing the voxel status and introducing faster ray-casting techniques. 
Schmid \textit{et al.} \cite{schmid2023dynablox} adopt Voxblox~\cite{oleynikova2017voxblox}, using the Truncated Signed Distance Function (TSDF) to construct the Euclidian Signed Distance Fields (ESDF).
These methods focus on free spaces but struggle with spaces where occupancy is uncertain due to the sparseness of rays, especially when used with low-cost sensors. 
While these algorithms attempt to extend the coverage of free cells by leveraging inflating and clustering, certain obstacle points remain unaddressed in the global map. 

\subsection{Differential-based Removal Algorithms}

Differential-based removal algorithms, unlike freespace-based methods, detect dynamic points based on visibility, reducing computational overhead compared to ray-casting methods. 
Pomerleau \textit{et al.} \cite{pomerleau2014long} use spherical coordinates for sparse point clouds and determine dynamic elements by dynamic probabilities of associated pairs of points. 
Kim \textit{et al.} \cite{kim2020remove} use a range image-based approach to separate static and dynamic points in urban environments. 
Others, like Lim \textit{et al.} \cite{lim2021erasor}, introduce height differences in volumes to incorporate cells and voxels. Recently, ERASOR2~\cite{lim2023erasor2} explore instant-aware mapping methods.
Although these methods improve computational efficiency in recognizing dynamic points, their map representation structures are egocentric, lacking global consistency, and necessitating additional transformations from global map points to the local frame.
Different from these methods, we perform difference detection in consistent global coordinates. Furthermore, we leverage binary encoding to enable fast matrix operations for the difference calculations.

\subsection{Data-driven Methods}
Mersch \textit{et al.}~\cite{mersch2022ral} employs sparse 4D convolutions to segment receding moving objects in 3D LiDAR data, efficiently processing spatiotemporal information using sparse convolutions. 
Sun \textit{et al.}~\cite{sun2022efficient} develop a novel framework for fusing spatial and temporal data from LiDAR sensors, leveraging range and residual images as input to the network. 
Pfreundschuh \textit{et al.} \cite{pfreundschuh2021dynamic} design an unsupervised network using an end-to-end occupancy grid-based approach for dynamic object labeling. 
Milioto \textit{et al.} \cite{milioto2019rangenet++} utilize range images combined with a neural network to address challenges in SqueezeNet \cite{wu2019squeezesegv2}, such as discretized displacement and blurred output. 
MotionSC \cite{wilson2022motionsc} develop a real-time dense local semantic mapping algorithm leveraging 3D deep learning for dynamic object supervision. 
These methods face challenges due to the types of labels and training models, which limit the accurate perception of only classified objects. 
Scene flow methods~\cite{fastflow3d,zhang2024deflow,zeroflow,li2023fast} provide a possible solution to estimate the velocity of points in a class-agnostic way. 
However, Domain adaption is still a crucial problem which means that the accuracy of methods decreased a lot if the train and test datasets were collected by different sensors~\cite{maciej2023applying}.

\section{Methodology}

\begin{figure*}[htbp]
  \centering
  \includegraphics[width=1.0\textwidth]{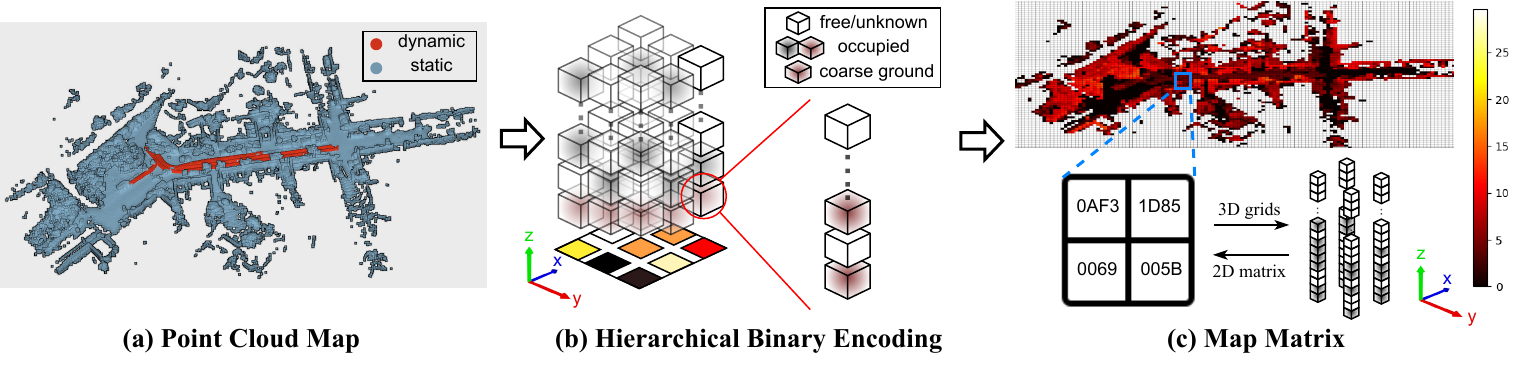}
  \caption{Implementation of our proposed map representation structure. (a) Global cluttered point cloud map. The red refers to dynamic points from ground truth only for visualization and no ground truth label will be used. (b) Map storage structure. Colored voxels mean they are occupied, and free voxels in white. Hierarchical resolutions on z-axis are applied to the ground grid cells shown in brown. (c) The binary-encoded 2D map matrix. The 2D matrix values of the zoom-in region are shown in hexadecimal, with the view of vertical occupancy data on the right. For a better view, each value is logarithmically mapped and visualized in heatmap format.}
  \label{fig:structure}
\end{figure*}
The BeautyMap framework is illustrated in \cref{fig:overview}. 
After turning point clouds into binary-encoded matrices (\cref{sec:binary_enc_ma}),
we mark potential dynamic regions by comparing matrices directly and use fine ground segmentation to extract dynamic parts near the ground (\cref{sec:dynamic_remove}). 
We then employ static restoration to protect points out of sight from being removed (\cref{sec:static_res}). We run our method on each scan independently and remove detected dynamic points in the end.
Further details are provided in the following.

\subsection{Binary-encoded Matrix}
\label{sec:binary_enc_ma}
\subsubsection{Matrix Generation}
The map representation approach is crucial for the dynamic points removal performance. 
Recent studies \cite{kim2020remove, schmid2023dynablox,wu2024moving} have extensively employed range image maps to enhance the computational efficiency of classic voxel-based ray-casting methods. 
However, their operation within the ego-centric frame lacks global consistency, which means a large number of global map points need to be converted into local frames. 
Our method uses consistent voxel-based structures for both scans and the global map to enhance efficiency, so that all processes can be performed within the same global frame. 

As shown in \cref{fig:structure}, our model retains a conventional grid structure. 
To facilitate efficient storage and retrieval, we compress the information into a 2D matrix, where each element is a binary number that contains the encoded vertical occupancy data. 
This structure necessitates only one-time processing of the global point cloud map, making the matrix immediately usable without any need for conversions. 

To prepare the data within a unified coordinate system for matrix indexing, we apply a transformation to the point cloud, followed by compression into a matrix $\mathbf{M} \in \mathbb{R}^{N_1 \times N_2}$. This reduces the complexity of subsequent operations to the order of the matrix. 

\begin{algorithm}
	\renewcommand{\algorithmicrequire}{\textbf{Input:}}
	\renewcommand{\algorithmicensure}{\textbf{Output:}}
	\caption{Binary-Encoding Algorithm}
	\label{alg:alg1}
	\begin{algorithmic}[1]
		\STATE Initialization: For each \(i, j\), set \(\mathbf{M}_{i,j} \leftarrow 0\).
            \STATE Set bit constraint \(N_{bit}\)
		\FOR{each point \(\mathbf{p} = \{p_x,p_y,p_z\}\) in \(pts\)}
    		\STATE \(i \leftarrow \text{floor}(p_x/\text{res}_g)\) 
    		\STATE \(j \leftarrow \text{floor}(p_y/\text{res}_g)\) 
    		\STATE \(k \leftarrow \text{floor}(p_z/\text{res}_h)\) 
                \IF{\(k < 0\)} 
                    \STATE Skip to the next iteration. 
                \ELSIF{\(k \geq N_{bit}\)}
                    \STATE \(\mathbf{M}_{i,j} \leftarrow \mathbf{M}_{i,j} \vee (1 \ll  (N_{bit}-1))\) 
                \ELSE
        		    \STATE \(\mathbf{M}_{i,j} \leftarrow \mathbf{M}_{i,j} \vee (1 \ll k)\) 
                \ENDIF
            \ENDFOR
		\ENSURE binary matrix $\mathbf{M}$
	\end{algorithmic}  
\end{algorithm}

\begin{figure}[t]
  \centering
  \includegraphics[width=\linewidth]{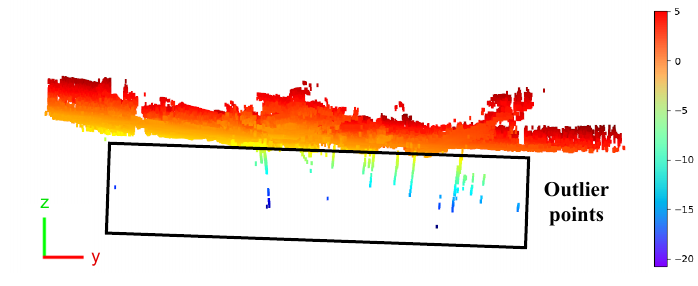}
  \caption{Underground outlier points in the black square are apparently shown in the side-view of the global point cloud map. The color bar indicates the point heights in meters.}
  \label{fig:MAD}
  \vspace{-1.0em} 
\end{figure}

We then encode the occupancy status of each vertical column of voxels bit-wise as a binary number. 
We define the status of the vertical cell as $1$ for \textit{occupied} and $0$ for others and perform bit operations based on the index subsets.
For instance, if a specific matrix element has points in its third upward cell, the cell occupancy value will be set as 1.
Any occupancy state beyond the maximum storable digits exceeding the height limit is considered part of the occupancy of the highest cell. 
The operation is also performed on each scan frame to obtain the query matrix within the unified coordinate system. 

\subsubsection{Coarse Ground Extraction}
\label{sec:corase_extraction}
During range sensor scanning, surfaces with high reflectivity, such as glass or water, can cause rays to have multiple reflections. 
This often leads to dense outlier point clusters below ground in certain regions as illustrated in the side-view in~\cref{fig:MAD}. 
These outliers will expand the boundaries and result in inefficient space utilization for traditional voxel structures and also affect the accuracy of the representation structure. 

To identify outlier points below ground and perform adaptable ground extraction, we employ the Median Absolute Deviation (MAD) algorithm~\cite{rousseeuw1993alternatives}. 
Superior in representing the central tendency and less influenced by outliers compared to other metrics, the MAD algorithm effectively identifies and handles outlier points in this situation. 
With the assumption that ground height changes are continuous, we determine ground height constraints by calculating the MAD value of the lowest points in a close range. 
Let $\mathcal{L}$ be the set of lowest point height values from the surroundings, then the process is represented by the following equations:
\begin{align}
    \text{MAD}(\mathcal{L}) &= \text{median}\left( |l_i - \text{median}(\mathcal{L})| \right) \ \forall l_i \in \mathcal{L} \\
    \{l_{\min}, l_{\max}\} &= \text{median}(\mathcal{L}) \pm 3 \times \text{MAD}(\mathcal{L}) \\
    g_k &= \min(\max(l_k, l_{\min}), l_{\max})
\end{align}
where $l_{k}$ is the lowest height value of the current vertical column, and $l_i$ is an individual height value from the set $\mathcal{L}$. $l_{\min}, l_{\max}$ is the lower and upper bound respectively.
We confine the ground height $g_k$ of the coarse ground cell $k$ to lie between $l_{\min}$ and $l_{\max}$. 
When encoding occupancy status into a 3D grid, $g_k$ serves as the starting height of the lowest cell in each vertical column. As shown in \cref{fig:structure}(b) and \cref{fig:LOGIC}(a), the 3d grid starts from different height cells, thereby better handling non-flat terrains.
For the global map, we introduce a hierarchical structure only for coarse ground cells.
To achieve the fine resolution structure, we encode points inside the ground cell into a 3D binary grid again as shown in \cref{fig:structure} (b).

\subsection{Dynamic Removal} 
\label{sec:dynamic_remove}
In this part, we first detect dynamic part radically by directly comparing map and query matrices from the previous steps.
Differences are marked as potential dynamic regions. 
Then, using the ground points in the static grid, we zoom in on the grid and use fine ground segmentation to further mark the missed dynamic points (see \cref{fig:LOGIC}).

\subsubsection{Matrix Comparison}
Based on our binary-encoded matrix-based map representation, we identify a matrix $\mathbf{M}_{dyn}$ indicating potential dynamic regions by calculating a difference matrix $\mathbf{M}_{diff}$ between each query scan matrix $\mathbf{M}_{scan}$ and the map $\mathbf{M}_{map}$ as following: 
\begin{align}
    \mathbf{M}_{dyn} &= \mathbf{M}_{map} \land (\neg \mathbf{M}_{scan}),
    \label{eq:comparsion}
\end{align}
where $\land$ denotes the AND operation and $\neg$ denotes the NOT operation. 

\subsubsection{Fine Ground Segmentation}
\label{sec:fine_ground}
Existing methods \cite{lim2021erasor, zhang2023dynamic, yan2023rh} utilize Principal Component Analysis or Sample Consensus segmentation to perform ground plane fitting, or estimate the normal direction to segment the ground plane. 
However, they still struggle with the considerable thickness of ground points in the global map which is caused by pose errors and severe terrain variations. 
The fixed resolution of these methods also makes it difficult to separate dynamic points close to the ground points distinctly.

In our approach, as detailed in \Cref{sec:corase_extraction}, we first define $g_k$ to identify coarse ground grid cells. 
These cells are initially determined based on their potential to be part of the ground. 
Since the cell with both dynamic points and the ground might be mistakenly detected as static, we introduce the concept of $C_{sel}$. 
A coarse ground grid cell is considered as $C_{sel}$ only if the adjacent cell above it is potentially dynamic.

\begin{figure}[t]
  \centering
  \includegraphics[width=\linewidth]{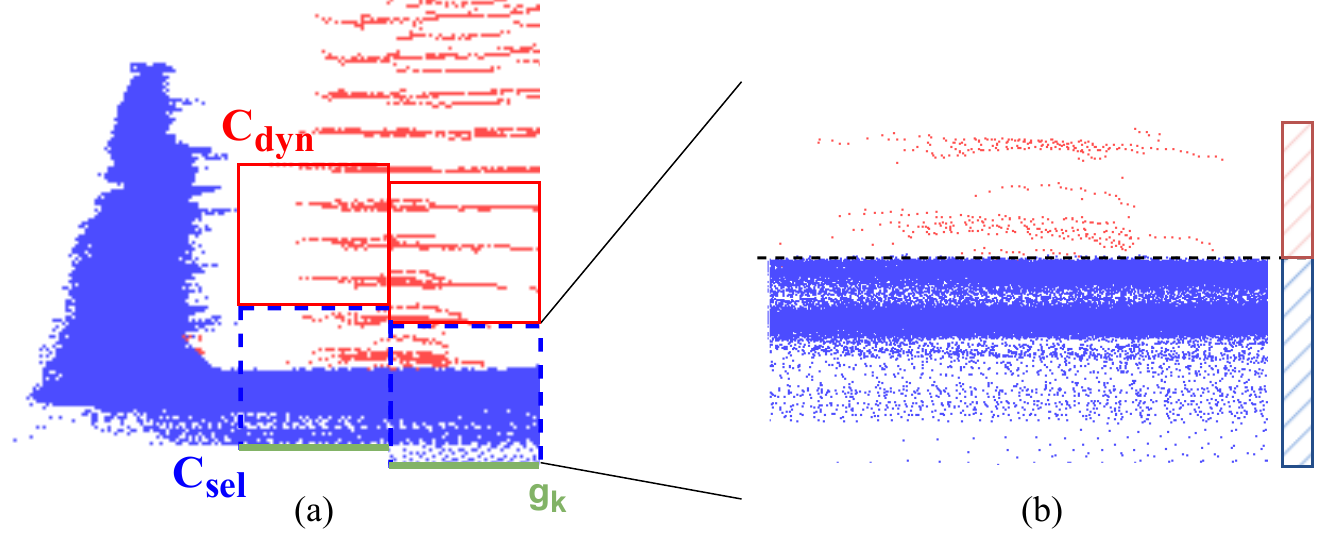}
  \caption{Our fine ground segmentation process. The black box zooms in on $C_{sel}$ showing the ground points number ratio segmentation along the black dashed line. \textcolor{blue}{Blue points} for static and \textcolor{red}{red points} for dynamic. The \textcolor{red}{red boxes} are potential dynamic grids and the dashed \textcolor{blue}{blue boxes} with both static and dynamic points are for internal calculation. }
  \label{fig:LOGIC}
\end{figure}

When building a global map, dynamic objects are moving, and ground points also accumulate through time. 
This results in different distribution of point numbers of the ground and dynamic points. 
Through this fact, we calculate the distribution of points in the coarse ground cell at higher resolution and apply a ground points number ratio (\cref{fig:LOGIC} (b)) to segment the ground that contains the most number of points. 

\begin{figure*}[t]
  \centering
  \includegraphics[width=\textwidth]{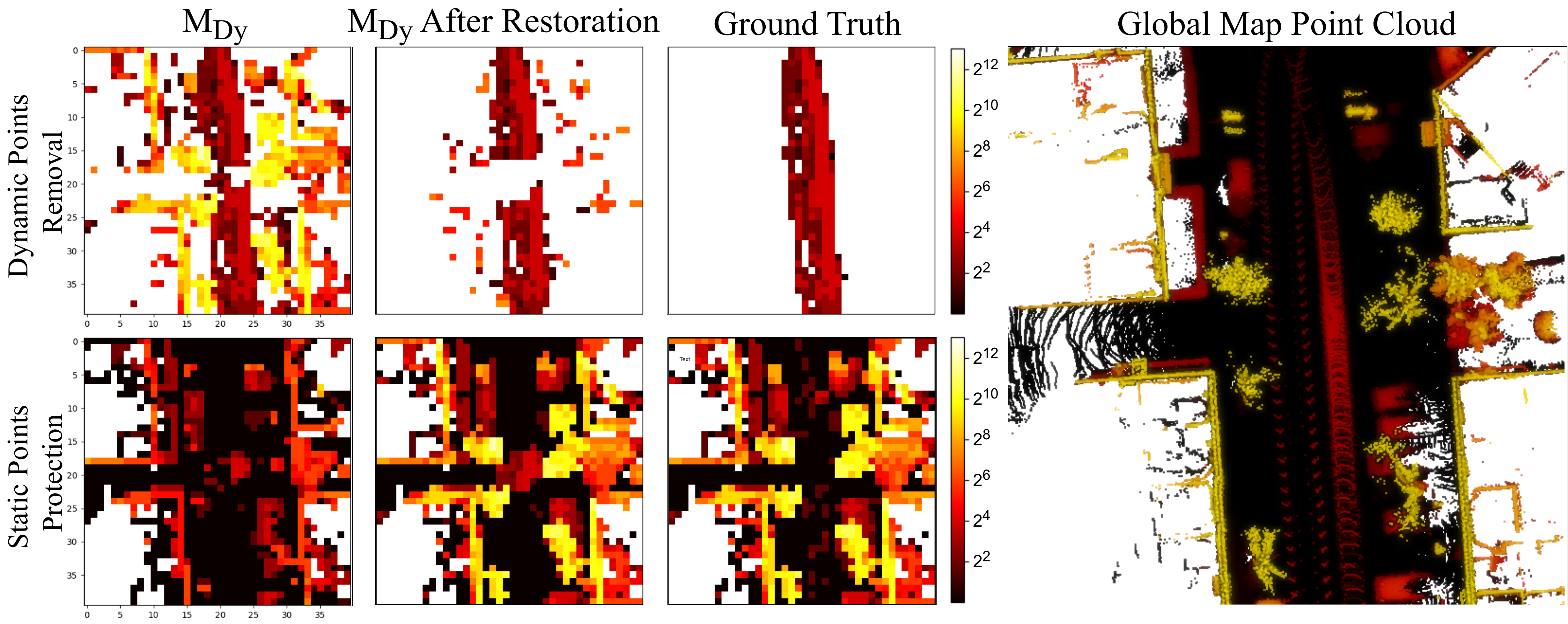}
  \caption{Visulization of matrices with potential dynamic regions. The first row shows the potential dynamic point region matrix, the dynamic region matrix after performing our static restoration method which has a noticeable decrease in the mistakenly identified static regions, and the ground truth. The second row displays the remaining static region matrix, the matrix after static restoration, and the ground truth. The original point cloud of the global map is shown on the right. }
  \label{fig:VVR}
\end{figure*}

\subsection{Static Restoration Module}
\label{sec:static_res}
Despite the ground segmentation, many static regions out of sight of the current scan are falsely included in $\mathbf {M}_{dyn}$ from \cref{eq:comparsion}.
Therefore, we implement further restoration strategies to balance dynamic point removal and static point retention. 
To restore these static points in the global map, we employ range and visibility-based methods according to the properties of sensors, producing mask matrices that refine our classification of dynamic regions. The results are shown in \cref{fig:VVR}.

Some points in the map may be out of sight in the current scan's view while inside the potential dynamic $\mathbf{M}_{dyn}$. 
Considering the LiDARs' field of view, we restore cells above the highest detected cells of each vertical column. 
The slope $k$ of the rays with the highest point is defined as: 
\begin{equation}
    k = \mathop{max}\limits_{i} (\frac{z_i}{\sqrt{x_i^2+y_i^2}}),
\end{equation}
where $i$ is the index of a point and $\mathbf{p}_i = (x_i, y_i, z_i)$ is the point's position. 
And the highest height of each column can be calculated in the global frame by $\lceil k\times s \rceil$, where $s$ is the 2D distance to the sensor. 
We then protect the higher regions from being classified as dynamic since they are outside the sight range.

\begin{figure}[t]
  \centering
  \includegraphics[width=0.48\textwidth]{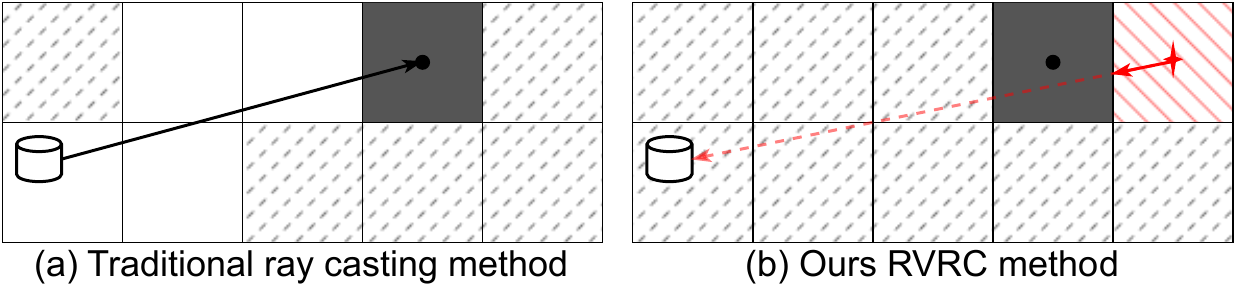}
  \caption{Comparison of traditional RC and ours RVRC. Gray dashes refer to unknown spaces, white shows the free spaces, black solid is the occupied cell, and red slash represents our potential dynamic cell.}
  \label{fig:RVRC}
  \vspace{-1em} 
\end{figure}

In dynamic removal tasks, ray casting means that the ray is projected from the sensor to detected points and helps identify free spaces along its path. 
Considering higher efficiency and effectiveness, we propose reverse virtual ray casting (RVRC). It casts virtual rays from each potential dynamic region back to the sensor. 
This reverse casting ends when meeting the first occupied cell along the virtual ray paths, indicating that this region is blocked and the potential dynamic cell is protected from being removed. 
Using grid-based ray casting methods, similar to accelerated methods \cite{duberg2020ufomap}, we ensure a balance between speed and accuracy.
We make it more efficient by only considering newly detected potential dynamic grids in subsequent scans within the global frame as shown in~\cref{fig:RVRC}.

\begin{table*}[htbp]
\caption{Quantitative comparison of dynamic points removal methods, $^\dagger$ means data-driven methods. The best results are shown in \textbf{bold} and the second best results are shown in \underline{underlined}. Results are in percentage.}
\centering
\def\arraystretch{1.2}
\scalebox{1.0}{
\begin{tabular}{l|ccc|ccc|ccc|ccc} 
\toprule
& \multicolumn{3}{c|}{KITTI sequence 00}
& \multicolumn{3}{c|}{KITTI sequence 01}
& \multicolumn{3}{c|}{KITTI sequence 05}
& \multicolumn{3}{c}{Semi-indoor}                                                                         \\ 
\hline
Methods & \multicolumn{1}{l}{SA ↑} & \multicolumn{1}{l}{DA ↑} & \multicolumn{1}{l|}{{\cellcolor[rgb]{0.949,0.949,0.949}}HA ↑}& \multicolumn{1}{l}{SA ↑} & \multicolumn{1}{l}{DA ↑}  & \multicolumn{1}{l|}{{\cellcolor[rgb]{0.949,0.949,0.949}}HA ↑} & \multicolumn{1}{l}{SA ↑} & \multicolumn{1}{l}{DA ↑}  & \multicolumn{1}{l|}{{\cellcolor[rgb]{0.949,0.949,0.949}}HA ↑} & \multicolumn{1}{l}{SA↑} & \multicolumn{1}{l}{DA ↑}  & \multicolumn{1}{l}{{\cellcolor[rgb]{0.949,0.949,0.949}}HA ↑}  \\ 
\hline
4DMOS$^\dagger$~\cite{mersch2022ral}  & - & - & {\cellcolor[rgb]{0.949,0.949,0.949}} - & - & - & {\cellcolor[rgb]{0.949,0.949,0.949}}- & - & - & {\cellcolor[rgb]{0.949,0.949,0.949}}- & 99.99 & 10.60 & {\cellcolor[rgb]{0.949,0.949,0.949}}27.59                     \\
MapMOS$^\dagger$~\cite{mersch2023ral}  & - & - & {\cellcolor[rgb]{0.949,0.949,0.949}} - & - & - & {\cellcolor[rgb]{0.949,0.949,0.949}}- & - & - & {\cellcolor[rgb]{0.949,0.949,0.949}}- & 99.99 & 4.75 & {\cellcolor[rgb]{0.949,0.949,0.949}}9.07 \\
DeFlow$^\dagger$~\cite{zhang2024deflow}  & 99.43 & 81.68 & {\cellcolor[rgb]{0.949,0.949,0.949}}89.69 & 99.19 & 81.25 & {\cellcolor[rgb]{0.949,0.949,0.949}}89.33 & 99.48 & 50.85 & {\cellcolor[rgb]{0.949,0.949,0.949}}67.30 & 99.99 & 2.02 & {\cellcolor[rgb]{0.949,0.949,0.949}}3.95                     \\
Removert~\cite{kim2020remove}  & 99.44                    & 41.53                   & {\cellcolor[rgb]{0.949,0.949,0.949}}58.59                     &  97.81                    & 39.56                    & {\cellcolor[rgb]{0.949,0.949,0.949}}56.33       &99.42                    & 22.28                              & {\cellcolor[rgb]{0.949,0.949,0.949}}36.40                                   & 99.96                    & 12.15                    & {\cellcolor[rgb]{0.949,0.949,0.949}}21.67                     \\
ERASOR~\cite{lim2021erasor}            & 66.70                    & 98.54                       & {\cellcolor[rgb]{0.949,0.949,0.949}}79.55                     &  98.12                    & 90.94                  & {\cellcolor[rgb]{0.949,0.949,0.949}}\underline{94.39}       & 69.40                    & 99.06                         & {\cellcolor[rgb]{0.949,0.949,0.949}}81.62                               & 94.90                    & 66.26                          & {\cellcolor[rgb]{0.949,0.949,0.949}}78.04                        \\
Octomap~\cite{hornung2013octomap}                        & 68.05                    & 99.69                        & {\cellcolor[rgb]{0.949,0.949,0.949}}80.89                         & 55.55                    & 99.60                   & {\cellcolor[rgb]{0.949,0.949,0.949}}71.28            & 66.28                    & 99.24                           & {\cellcolor[rgb]{0.949,0.949,0.949}}79.48                     & 88.97                    & 82.18                    & {\cellcolor[rgb]{0.949,0.949,0.949}}\underline{85.44}                   \\
Octomap w GF ~\cite{zhang2023dynamic}                        & 93.06                    & 98.67                     & {\cellcolor[rgb]{0.949,0.949,0.949}}\underline{95.78}                     &  80.64                    & 97.27                       & {\cellcolor[rgb]{0.949,0.949,0.949}}88.18               & 93.54                    & 92.48              & {\cellcolor[rgb]{0.949,0.949,0.949}}93.01                     &96.79                    & 73.50                                & {\cellcolor[rgb]{0.949,0.949,0.949}}83.55             \\
dynablox~\cite{schmid2023dynablox}                                         & 96.76                    &90.68                     & {\cellcolor[rgb]{0.949,0.949,0.949}}93.62                                       & 96.33                    & 68.01                          & {\cellcolor[rgb]{0.949,0.949,0.949}}79.73          & 97.80                    & 88.68              & {\cellcolor[rgb]{0.949,0.949,0.949}}\underline{93.02}               & 98.81  & 36.49                          &{\cellcolor[rgb]{0.949,0.949,0.949}}53.30                     \\
BeautyMap (Ours)                                        & 96.76                    &98.38                              & {\cellcolor[rgb]{0.949,0.949,0.949}}\bf{97.56}                 & 99.17                    & 92.99                 & {\cellcolor[rgb]{0.949,0.949,0.949}}\bf{95.98}      & 96.34                    & 98.29                       & {\cellcolor[rgb]{0.949,0.949,0.949}}\bf{97.31}                   & 93.69                    & 90.67                        & {\cellcolor[rgb]{0.949,0.949,0.949}}\bf{92.16}                  \\

\bottomrule
\end{tabular}
}
\label{num_table}
\end{table*}
\section{Experiment}

We follow evaluation protocol from~\cite{zhang2023dynamic}, the experiments dataset of different sensor types includes KITTI~\cite{Geiger2013IJRR} (VLP-64) where dynamic ground truths are from Semantic-Kitti~\cite{behley2019semantickitti} and semi-indoor~\cite{zhang2023dynamic} datasets (VLP-16). 
We kept the same setting from \cite{zhang2023dynamic,lim2021erasor} that selected partial frames from the KITTI sequences 00, 01, 02, and 05 which present the most dynamics.
All poses of datasets are from SLAM packages like SuMa~\cite{behley2018rss} for KITTI and simple-ndt for semi-indoor~\cite{simple-ndt-slam}.
The metrics of dynamic removal in maps are Static Accuracy (SA), and Dynamic Accuracy (DA) at the point level without downsampling the ground truth map to have an accurate and fair evaluation. Instead of using Associate Accuracy (AA) proposed in~\cite{zhang2023dynamic}, we propose to
use the Harmonic Accuracy (HA) as a comprehensive metric that combines both accuracies.
    \begin{equation*}
        HA = \frac{2\times SA \times DA}{SA+DA}
    \end{equation*}

Beside methods provided in the public benchmark~\cite{zhang2023dynamic}, we add three more data-driven methods: 4DMOS~\cite{mersch2022ral}, MapMOS~\cite{mersch2023ral} and DeFlow~\cite{zhang2024deflow} into our comparison. The training dataset of 4DMOS and MapMOS includes KITTI sequences from 0 to 10. No results are therefore presented for the KITTI sequences as these methods have seen ground truth data.
DeFlow is trained on the Argoverse 2 \textit{Sensor} dataset. 
The default parameters for BeautyMap are $1\times1$~m${}^2$ cell size for all KITTI sequences and $0.5\times0.5$~m${}^2$ cell size for the semi-indoor dataset. \Cref{tab:voxel_runtime} shows an ablation study on the tradeoff between different cell sizes, performance, and running speed.

The computation cost is also compared between our binary-encoded structure-based method and others.
All experiments are conducted on a desktop computer equipped with an Intel Core i9-12900KF processor.

\subsection{Quantitative Results}
\begin{figure*}[t]
  \centering
  \includegraphics[width=\linewidth]{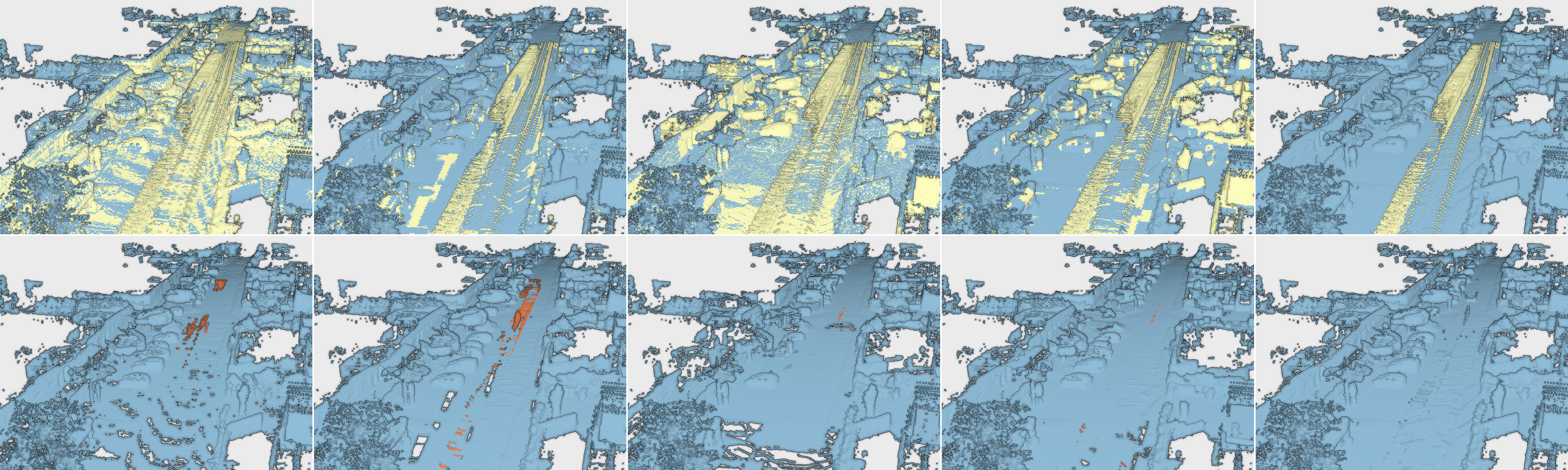}
  \newcolumntype{C}{>{\centering\arraybackslash}X}
\begin{tabularx}{\linewidth}{@{} *{5}{C} @{}}
  \small (a) Octomap\textsuperscript~\cite{hornung2013octomap} & \small (b) dynablox~\cite{schmid2023dynablox} & \small \hspace{-15pt}(c) ERASOR~\cite{lim2021erasor} & \small \hspace{-5pt}(d) BeautyMap (Ours) & \small (e) Ground Truth \\[0pt]
\end{tabularx}
  \caption{\small Qualitative results in KITTI Sequence 05 using one VLPC-64 LiDAR sensor. 
  The first row represents the labeling result. The second row shows the cleaned output map from different approaches. In all the methods shown in the images, \colorbox[HTML]{FFFFBF}{\textcolor{black}{yellow}} represents points labeled as dynamic, and \colorbox[HTML]{ef8554}{\textcolor{black}{orange}} indicates dynamic points that are incorrectly labeled as static. 
  }
  \label{fig:full_av}
  \vspace{-0.5em}
\end{figure*}

The quantitative comparison results of dynamic points removal methods on several datasets are shown in~\cref{num_table}. 
Regarding differential-based methods, ERASOR shows its ability to remove dynamic points and achieve a high DA score, at the cost of reducing the SA score. Removert shows an opposite low score, which is mainly attributed to the attendance of incorrectly recovered dynamic points near the ground with the coarse range image, as mentioned in \cite{kim2020remove}. 
The dynamic points removal results of the semi-indoor dataset on 4DMOS, MapMOS and DeFlow reveal that these three data-driven methods struggle with 16-channel LiDAR. Since both are trained on datasets captured with LiDARs with many channels like KITTI (64-channel) and AV2 (32-channel), one possible way is to label and train on mixed datasets.
The freespace-based method Octomap struggles to balance static point saving and dynamic removal because of noise and ground point as mentioned in \cite{zhang2023dynamic}. The other two, Octomap w GF and dynablox, usually maintain good static and dynamic accuracy.
The special case is on the semi-indoor dataset (16-channel), where 4DMOS, MapMOS, DeFlow, Removert, and dynablox perform poorly on dynamic accuracy. The reason behind it for Removert and dynablox is that they transfer LiDAR points to a range image and the resolution of this range image for LiDARs with few channels limits the detection ability. 
As a differential-based method, our method benefits from the new point cloud representation and the difference strategy we propose and outperforms all other methods in all four datasets in~\cref{num_table}. 

\subsection{Qualitative Results}
The qualitative results for dynamic removal in point cloud maps on KITTI sequence 05 are presented in \cref{fig:full_av}. The first row displays the labeling results, while the second row shows the cleaned output maps, with remaining dynamic points marked in orange.
For freespace-based methods, Octomap (\cref{fig:full_av}(a)) exhibits a considerable number of false dynamic labels in the first row and misclassified dynamic points in the cleaned map in the second row. Dynablox (\cref{fig:full_av}(b)), on the other hand, is more conservative in identifying dynamic grids, as evident from the results in the first row.
ERASOR (\cref{fig:full_av}(c)) produces a cleaner map compared to the other two methods, but it also detects more false dynamic points and inaccurately removes some static structures.
The cleaned map of our BeautyMap most closely matches the ground truth, outperforming the other methods in terms of accuracy and dynamic point removal.

\subsection{Computation Cost Comparison}
We compare the runtime per frame of our proposed BeautyMap method with other dynamic point removal methods on KITTI sequence 01.  
As shown in \cref{tab:runtime_comparison}, our BeautyMap method outperforms the others, performing a runtime of 0.046 seconds per point cloud. 
This efficiency is attributed to our binary-encoded matrix structure. 
The second fastest method Removert~\cite{kim2020remove} utilizes range image and direct comparison between the map and scans to realize a shorter runtime. 
The Dynablox \cite{schmid2023dynablox} has its runtime close to that of the Removert method, with the TSDF structure to identify the free spaces and decrease the complexity. 
The ERASOR method \cite{lim2021erasor} follows the rank by encoding the point clouds into Volume of Interest (VOI) and performing bin-wise calculations based on this structure. 
The Octomap \cite{hornung2013octomap} and Octomap w GF \cite{zhang2023dynamic}, using traditional Octree structure and ray casting method for each point, are the slowest. 

In exploring the tradeoffs between cell size, speed, and performance in our BeautyMap method, we present a comparison of different cell sizes in~\cref{tab:voxel_runtime} using KITTI sequence 02. As shown in~\cref{tab:voxel_runtime}, a smaller cell size (0.5 meters) leads to a slight increase in HA, the improvement is marginal compared to the significant increase in runtime. On the other hand, a larger cell size (2.0 meters) results in higher DA but lower SA, as it tends to remove more points. This default setting also offers the best runtime efficiency and well dynamic points removal performance. Comparing the results for BeutyMap in \cref{tab:runtime_comparison} and \cref{tab:voxel_runtime} with 1.0 meter cell size it can be seen that the runtime is consistently low in both datasets.

\begin{table}[t]
\centering
\def\arraystretch{1.2}
\caption{Runtime comparison of different methods running on KITTI sequence 01.}
\begin{tabular}{lc}
\toprule
Methods      & Runtime/point cloud [s] \(\downarrow\)\\
\midrule
Removert~\cite{kim2020remove}       & \underline{0.134} $\pm$ 0.004      \\
ERASOR~\cite{lim2021erasor}         & 0.718 $\pm$ 0.039      \\
Octomap~\cite{hornung2013octomap}   & 2.981 $\pm$ 0.952       \\
Octomap w GF~\cite{zhang2023dynamic} & 2.147 $\pm$ 0.468      \\
Dynablox~\cite{schmid2023dynablox}  & 0.141 $\pm$ 0.022   \\
BeautyMap (Python)                 & \textbf{0.046} $\pm$ 0.011        \\
\bottomrule
\end{tabular}
\label{tab:runtime_comparison}
\end{table}

\begin{table}[h]
\centering
\def\arraystretch{1.2}
\caption{Runtime comparison of BeautyMap with different cell size on KITTI sequence 02. Runtime is on per point cloud.}
\begin{tabular}{ccccc}
\toprule
Size [\unit{\metre}] & SA $\uparrow$ & DA $\uparrow$ & HA $\uparrow$ & Runtime/point cloud [s]  \\
\midrule
0.5  & 83.92 & 84.14 & \textbf{84.03} & 0.132 $\pm$ 0.034      \\
1.0  & 83.40 & 82.41 & 82.90 & 0.031 $\pm$ 0.039      \\
2.0  & 74.92 & 88.83 & 81.28   & \textbf{0.018 $\pm$ 0.010}       \\
\bottomrule
\end{tabular}
\label{tab:voxel_runtime}
\end{table}

\subsection{Ablation Study}

In our proposed method, matrix comparison is the main module to classify potential dynamic points. But other modules are equally important to improve the performance. Therefore, we carry out an ablation study on KITTI sequence 01 to further evaluate the contributions of adaptable ground adjustment modules and static restoration. 

When performing matrix comparison without any other modules, our method achieves a relatively high DA score but a low SA score as shown in \cref{tab:ablation_study}. This kind of map results in losing many static features and may harm localization accuracy. 

The ground module further includes dynamic points that are close to the ground proposed in \cref{sec:fine_ground}. 
We achieve higher dynamic accuracy with the ground module and static accuracy is still similar to the previous one, proving that points we remove mostly belong to dynamic objects.

However, both of them have relatively low static accuracy. That's where static restoration can help with protecting the out of sight points from being removed. With static restoration, we observe that SA is higher around 40\% than the other two and near 100\%.

The complete BeautyMap method with ground and static restoration modules together can achieve state-of-the-art results in dynamic point removal in a map. More than 90\% points in the map are correctly classified and output a satisfied cleaned map with HA to 96\%.

\begin{table}[t]
\caption{Ablation Study of Function Modules} 
\label{tab:ablation_study}
\centering
\def\arraystretch{1.3}
\begin{tabular}{c c c c c}
\toprule
Ground module & Static restoration & SA $\uparrow$ & DA $\uparrow$  & {\cellcolor[rgb]{0.949,0.949,0.949}} HA $\uparrow$ \\
\hline
   &   & 59.55 & 96.47 & {\cellcolor[rgb]{0.949,0.949,0.949}}75.79\\
  \checkmark &   &  59.04 & \textbf{99.01} & {\cellcolor[rgb]{0.949,0.949,0.949}}76.46 \\
    & \checkmark & \textbf{99.38} & 77.81 & {\cellcolor[rgb]{0.949,0.949,0.949}}87.94 \\
   \checkmark & \checkmark   & 99.17 & 92.99 & {\cellcolor[rgb]{0.949,0.949,0.949}}\textbf{96.03} \\ 
\bottomrule
\end{tabular}
\end{table}

\section{Conclusions}
In this paper, we introduce `BeautyMap', a novel methodology for dynamic points removal in global point cloud maps. 
Our approach leverages the binary-encoded matrix map representation structure, together with adaptable ground adjustment and static restoration modules, to become a considerable solution for global map maintenance. 

Through comparative experiments, BeautyMap reveals its superior capabilities in both accuracy and efficiency against other traditional methods, showing the efficiency of our structure and methodology. 
It can also be generalized for diverse scenarios for dynamic points removal and is suitable for range sensors of various types. 

Like many other methods, BeautyMap also faces the challenge of dealing with multiple ground levels at a specific position. A potential solution for our proposed method is to detect the ceiling and perform better protection.

In future works, we will explore further usage of the binary-encoded matrix structure, 
deal with challenging scenes like multiple ground levels, implement an online mode integrated with real-time odometry,
and expand our methodology for long-term change detection and map update frameworks.

\bibliographystyle{IEEEtran}
\bibliography{IEEEabrv,ref}
\end{document}